\documentclass[sigconf]{acmart}

\renewcommand\footnotetextcopyrightpermission[1]{} 
\usepackage{algorithm}
\usepackage{algorithmicx}
\usepackage{graphicx}
\usepackage{multirow}
\usepackage{caption}
\usepackage{subcaption}
\usepackage{xtab,booktabs}

\begin{document}
\fancyhead{}

\title{ThumbNet: One Thumbnail Image Contains All You Need for Recognition}

\author{Chen Zhao}
\email{chen.zhao@kaust.edu.sa}
\affiliation{%
  \institution{King Abdullah University of Science and Technology (KAUST), Saudi Arabia}
}

\author{Bernard Ghanem}
\email{bernard.ghanem@kaust.edu.sa}
\affiliation{%
  \institution{King Abdullah University of Science and Technology (KAUST), Saudi Arabia}
}

\begin{abstract}
Although deep convolutional neural networks (CNNs) have achieved great success in computer vision tasks, its real-world application is still impeded by its voracious demand of computational resources. Current works mostly seek to compress the network by reducing its parameters or parameter-incurred computation, neglecting the influence of the input image on the system complexity. Based on the fact that input images of a CNN contain substantial redundancy, in this paper, we propose a unified framework, dubbed as ThumbNet, to simultaneously accelerate and compress CNN models by enabling them to infer on one thumbnail image. We provide three effective strategies to train ThumbNet. In doing so, ThumbNet learns an inference network that performs equally well on small images as the original-input network on large images. With ThumbNet, not only do we obtain the thumbnail-input inference network that can drastically reduce computation and memory requirements, but also we obtain an image downscaler that can generate thumbnail images for generic classification tasks. Extensive experiments show the effectiveness of ThumbNet, and demonstrate that the thumbnail-input inference network learned by ThumbNet can adequately retain the accuracy of the original-input network even when the input images are downscaled 16 times.

\end{abstract}


\maketitle

\section{Introduction}\label{sec:intro}


%

 Recent years have witnessed not only the growing performance of deep convolutional neural networks (CNNs) \cite{simonyan2014very, he2016deep, xie2017aggregated, zhao2018boostnet, xu2020g, zhao2017better, zhao2018cream}, but also their expanding computation and memory costs \cite{canziani2016analysis}. Though the intensive computation and gigantic resource requirements are somewhat tolerable in the training phase thanks to the powerful hardware accelerators (e.g. GPUs), when deployed in real-world systems, a deep model can easily exceed the computing limit of hardware devices. Mobile phones and tablets, which have constrained power supply and computational capability, are almost intractable to run deep networks in real-time. A cloud service system, which needs to respond to thousands of users, has an even more stringent requirement of computing latency and memory. Therefore, it is of practical significance to accelerate and compress CNNs for test-time deployment.

\setlength\belowcaptionskip{-1ex}
\begin{figure}[t]
\begin{center}
\includegraphics[width=0.47\textwidth]{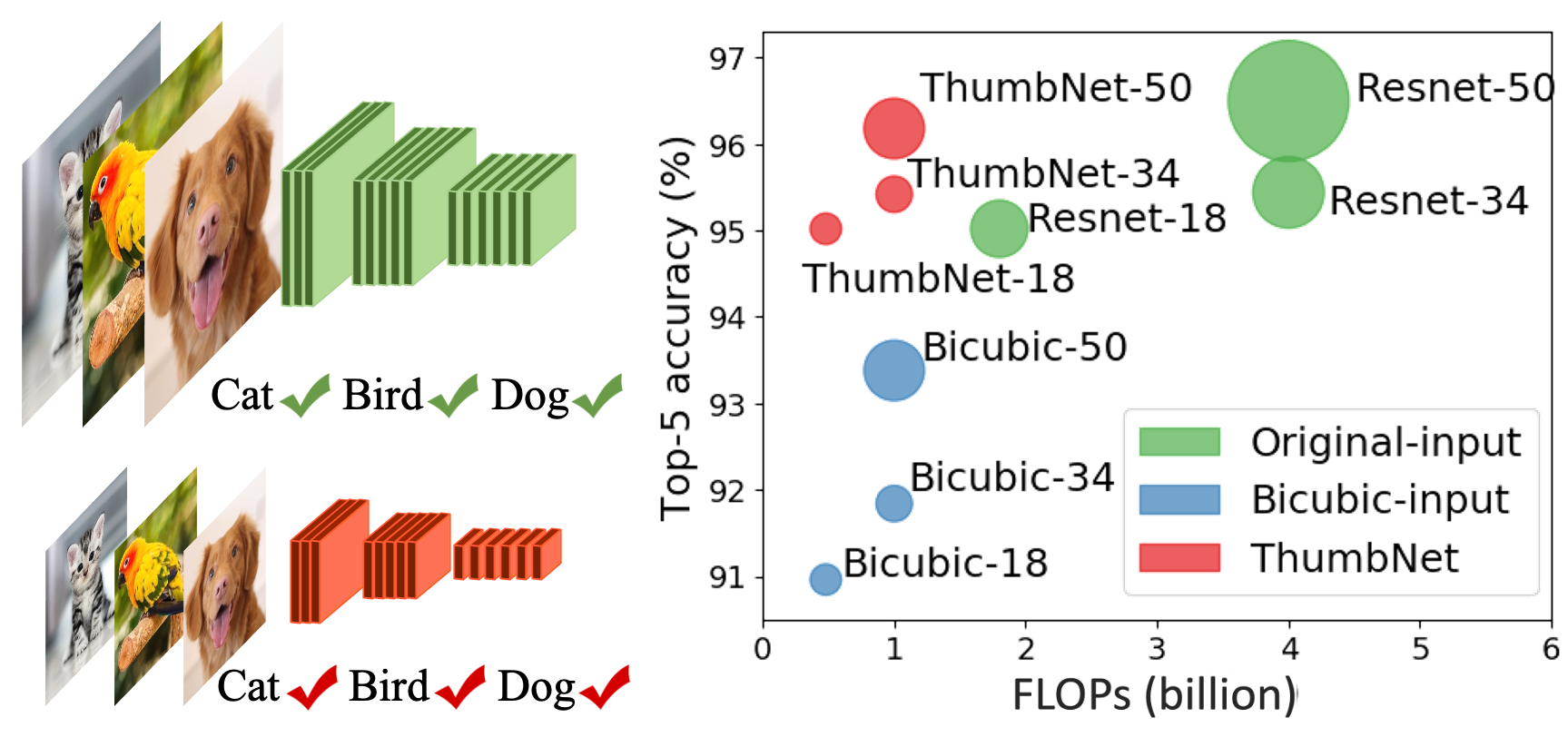}
\end{center}
\caption{\small{\textbf{Same CNN with different input sizes}. We accelerate a CNN by inferring on thumbnail images. Compared to the original-input network shown on the top-left, the thumbnail-input CNN shown on the bottom-left has the same architecture but smaller feature maps in all convolutional layers, hereby tremendously reducing computation and memory requirements, as shown on the right (circle sizes indicate memory requirements). The proposed ThumbNet can well retain the accuracy of the original-input networks, significantly outperforming the bicubic-input networks that input small images downscaled via bicubic interpolation.}}
\label{fig:motivation}
\end{figure}

Before delving into the question of how to speed up deep networks, let us first analyze what dominates the computational complexity of a CNN. We calculate the total time complexity of convolutional layers as follows \cite{he2015convolutional}:
\setlength{\abovedisplayskip}{5pt}
\setlength{\belowdisplayskip}{5pt}
\begin{equation}\label{eq:complexity}
\mathcal{O} \left( \sum \nolimits_{l=1}^d n_{l-1} \cdot s^2_l \cdot n_l \cdot m^2_l\right),
\end{equation}
where $l$ is the index of a layer and $d$ is the total number of layers; $n_{l-1}$ is the number of input channels and $n_{l}$ is the number of filters in the $l$-th layer; $s_l^2$ is the spatial size of the filters and $m_l^2$ is the spatial size of the output feature map.

Decreasing any factor in Eq. (\ref{eq:complexity}) can lead to reduction of total computation. One way is to sparsify network parameters by filter pruning \cite{lecun1990optimal, han2015deep}, which by defining some mechanism to prioritize the parameters, sets unimportant ones to zero. However, some researchers claim that these methods usually require sparse BLAS libraries or even specialized hardware. Hereby, they propose to prune filters as a whole \cite{li2016pruning, Luo2017ThiNetAF}. Another method is to decrease the number of filters by low rank factorization \cite{denton2014exploiting, jaderberg2014speeding, lebedev2014speeding, zhang2016accelerating}. If a more dramatic change in network structure is required, knowledge distillation \cite{hinton2015distilling, romero2014fitnets} can do the trick. It generates a new network, which can be narrower (with fewer filters) or shallower (with fewer layers), by transferring hidden information from the original network. Moreover, there are other approaches to lower the convolution overload by means of fast convolution techniques (e.g. FFT \cite{mathieu2013fast} and the Winograd algorithm \cite{lavin2016fast}), and quantization \cite{han2015deep, wu2016quantized} and binarization \cite{courbariaux2016binarized, rastegari2016xnor}.

All the above methods attempt to accelerate or compress neural networks from the viewpoint of network parameters, thereby neglecting the significant role that the spatial size of feature maps are playing in the overall complexity. According to Eq. (\ref{eq:complexity}), the required computation is diminished as the spatial size of feature maps decreases. Moreover, the memory required to accommodate those feature maps at run-time will also be reduced. Given a CNN architecture, we can simply decrease the spatial size of all feature maps by reducing the size of the input image.

In this paper, we propose to use a thumbnail image, i.e., an image of lower spatial resolution than its original-size counterpart, as test-time network input to accelerate and compress CNNs of any architecture and of any depth and width. This thumbnail-input network can dramatically reduce computation as well as memory requirements, as shown in Fig. \ref{fig:motivation}.

\vspace{4pt}\noindent \textbf{Contributions.} \textbf{(1)} We propose an orthogonal mechanism to accelerate a deep network compared to conventional methods: from the novel perspective of enabling  the  network to infer  on \textit{one single} downscaled  image efficiently and effectively.  To this end, we propose a unified framework called ThumbNet to train a thumbnail-input network that can tremendously reduce computation and memory consumption while maintaining the accuracy of the original-input network. \textbf{(2)} We present a supervised image downscaler that generates a thumbnail image with good discriminative properties and a natural chromatic look. This downscaler is reliably trained by exploiting \textit{supervised image downscaling}, \textit{distillation-boosted supervision}, and \textit{feature-mapping regularization}. The ThumbNet generated images can replace their original-size counterparts and be stored for other classification-related tasks, reducing resource requirements in the long run. \textbf{(3)} The proposed ThumbNet effectively preserves network accuracy at speedup ratios of up to $4\times$ (Imagenet) and $16\times$ (Places) on various networks, surpassing other network acceleration/compression methods by significant margins.

\section{Related Work}\label{sec:related}

In this section, we give an overview of the related works to our proposed ThumbNet in the literature.

\subsection{Knowledge Distillation}

Knowledge Distillation (KD) \cite{hinton2015distilling} was introduced as a model compression framework, which aims to reduce the computational complexity of a deep neural network by transferring knowledge from its original architecture (teacher) to a smaller one (student). The student is penalized according to the discrepancy between the softened versions of the teacher's and student's output logits\footnote{In this paper, we use `logits' to refer to the output of a neural network before the softmax activation function in the end.}. It claims that this teacher-student paradigm easily transfers the generalization capability of the teacher network to the student network in that the student not only learns the characteristics of the correct labels but can also benefit from the invisible finer structure in the wrong labels. There are some extensions to this work, e.g., using intermediate representations as hints to train a thin-and-deep student network \cite{romero2014fitnets}, applying it in object detection models \cite{chen2017learning}, and using it to enhance network resilience to adversarial samples \cite{papernot2016distillation}. These works mostly focus on learning a new network architecture. In our paper, we utilize the idea of KD to train the same network architecture with thumbnail images as input.

\subsection{Auto-Encoder}

An auto-encoder \cite{hinton2006reducing} is an unsupervised neural network, which learns a data representation of reduced dimensions by minimizing the difference between input and output. It consists of two parts, an encoder which maps the input to a latent feature, and a decoder which reconstructs the input from the latent feature. Early auto-encoders are mostly composed of  fully-connected layers \cite{hinton2006reducing, bourlard1988auto}. These days, with the popularity of CNNs, some researchers propose to incorporate convolution to an auto-encoder and design a convolutional auto-encoder \cite{turchenko2017deep, masci2011stacked}, which utilizes convolution / pooling to downscale the image in the encoder and utilizes deconvolution \cite{zeiler2010deconvolutional} / unpooling in the decoder to restore the original image size. Though the downscaled images from the encoder are effective for reconstructing the original images, they do not perform well for classification due to lack of supervision. In our work, instead of using convolutional auto-encoder as a downscaler, we incorporate it into ThumbNet as unsupervised pre-training to regularize the classification task.

\subsection{Downscaled Image Representation}

Representing an image with a downscaled size is an effective way to reduce computational complexity. One emerging method is to apply compressive sensing \cite{zhao2017better, zhao2018cream, Zhang2012ImageCS, Zhang2014ImageCS, Zhao2017VideoCS, Zhao2016NonconvexLN, Zhao2014ImageCR} when acquiring an image to sample only a small number of measurements. But these measurements are not in the format of 2D grid and cannot be directly processed by a CNN. Alternatively, the work WIDIC \cite{zhao2013wavelet} proposes to downscale an image into a smaller image in the wavelet domain, so as to increase its coding efficiency without compromising reconstruction quality. The proposed ThumbNet also preserves the image grid when downscaling but in a learnable manner. It focuses on the classification task and aims to increase network efficiency instead. For the task of classification, we find one recent work \cite{chen2018learning} (denoted here as LWAE), which also attempts to accelerate a neural network by using small images as input. It decomposes the original input image into two low-resolution sub-images, one with low frequency which is fed into a standard classification network, and one with high frequency which is fused with features from the low-frequency channel by a lightweight network to obtain the classification results. Compared to LWAE, our ThumbNet is able to achieve higher network accuracy with \emph{one single} thumbnail-image, resulting in fewer requirements for computation, memory and storage.


\begin{figure*}[htbp]
\begin{center}
\includegraphics[width=1.0\textwidth]{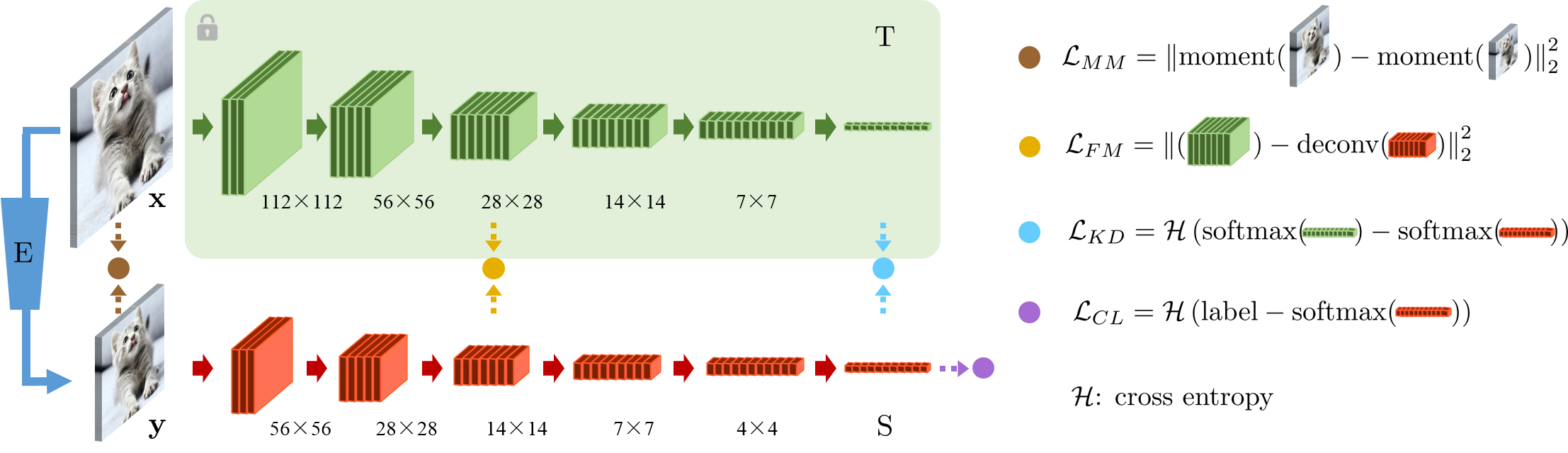}
\end{center}
\caption{\small{\textbf{ThumbNet Architecture}. Green: well-trained network $\mathrm{T}$; red: inference network $\mathrm{S}$; blue: downscaler $\mathrm{E}$. Blocks represent feature maps, solid arrows contain network operations such as convolution, the rectified linear unit (ReLU) \cite{nair2010rectified}, pooling, batch normalization \cite{ioffe2015batch}, etc. The numbers below each feature map are their spatial resolutions. Dots represent the four losses, which are moment-matching (MM) loss in brown, feature-mapping (FM) loss in yellow, knowledge-distillation (KD) loss in cyan and classification (CL) loss in purple. }}
\label{fig:architecture_proposed}
\end{figure*}

\section{Proposed ThumbNet}\label{sec:method}

\subsection{Network Architecture}

We illustrate the architecture of ThumbNet in Fig. \ref{fig:architecture_proposed}. The well-trained network $\mathrm{T}$ takes an input image of a large size, e.g., $224\times 224$, passes it through stacked convolutional layers and fully-connected layers, and produces $K$ logits, where $K$ is the number of classes. Its well-trained parameters $\mathbf{W}_{\mathrm{T}}$, are not changed during the whole training process of ThumbNet and only provide guidance for the inference network. The inference network $\mathrm{S}$, which takes as input an image of a small size, e.g., $112\times 112$. Each layer of $\mathrm{S}$ (except for the first fully-connected layer if its size is influenced by the input image size as in VGG), has exactly the same shape and size as its corresponding layer in $\mathrm{T}$. Each feature map of $\mathrm{S}$ has the same number of channels as its corresponding feature map of $\mathrm{T}$ but is smaller in the spatial size. The parameters in $\mathrm{S}$, denoted as $\mathbf{W}_{\mathrm{S}}$, are the main learning objectives of ThumbNet. The downscaler $\mathrm{E}$ generates a thumbnail image from the original input image, whose parameters are denoted as $\mathbf{W}_{\mathrm{E}}$.

\subsection{Details of Network Design}

There are three main techniques in the proposed ThumbNet: 1) {supervised image downscaling}, 2) {distillation-boosted supervision}, and 3) {feature-mapping regularization}.
\label{sec:method_details}

\subsubsection{Supervised Image Downscaling}

 Traditional image downscaling methods (e.g., bilinear and bicubic \cite{lehmann1999survey}) do not consider the discriminative capability of the downscaled images, which as a consequence lose critical information for classification. We instead exploit CNNs to adaptively extract discriminative information from the original images to tailor to the classification goal.

\begin{algorithm*}[htbp]
\caption{\small{\textbf{Training Strategy of ThumbNet}}
\small
\newline The well trained parameters $\mathbf{W}_{\mathrm{T}}$ of the original network are provided as input to the algorithm. All the trainable parameters are initialized as random values which are denoted by $\mathbf{W}_{\mathrm{S}}^0$, $\mathbf{W}_{\mathrm{E}}^0$, $\mathbf{W}_{\mathrm{D}}^0$.} \label{alg}
\begin{algorithmic}[1]
\State \textbf{Input:} $\mathbf{W}_{\mathrm{T}}$, $\mathbf{W}_{\mathrm{S}}^0$, $\mathbf{W}_{\mathrm{E}}^0$, $\mathbf{W}_{\mathrm{D}}^0$
\State $\mathbf{W}_{\mathrm{E}}^\ast, \mathbf{W}^\ast_{\mathrm{S}_l}, \mathbf{W}^\ast_{\mathrm{D}} \; \gets  \underset{\mathbf{W}_{\mathrm{E}}, \mathbf{W}_{\mathrm{S}_l}, \mathbf{W}_{\mathrm{D}} }{\operatorname{arg\,min}}    \mathcal{L}_{MM} \left(\mathbf{W}_{\mathrm{E}}\right) + \alpha \mathcal{L}_{FM}\left(\mathbf{W}_{\mathrm{E}}, \mathbf{W}_{\mathrm{S}_l}, \mathbf{W}_{\mathrm{D}}\right) + \frac{1}{2} \theta \mathcal{R}_{\mathrm{E}, \mathrm{S}_l, \mathrm{D}}$
\State $\mathbf{W}_{\mathrm{E}}^\ast, \mathbf{W}_{\mathrm{S}}^\ast  \; \gets  \underset{\mathbf{W}_{\mathrm{E}}, \mathbf{W}_{\mathrm{S}}}{\operatorname{arg\,min}} \;  \mathcal{L}_{CL} \left(\mathbf{W}_{\mathrm{E}}, \mathbf{W}_{\mathrm{S}}\right) + \beta \mathcal{L}_{KD} \left(\mathbf{W}_{\mathrm{E}}, \mathbf{W}_{\mathrm{S}}\right) + \frac{1}{2} \theta \mathcal{R}_{\mathrm{E}, \mathrm{S}}$
\State \textbf{Output:} $\mathbf{W}_{\mathrm{E}}^\ast$, $\mathbf{W}_{\mathrm{S}}^\ast$
\end{algorithmic}
\end{algorithm*}

For the sake of computational efficiency and simplicity, our supervised image downscaler $\mathrm{E}$ merely comprises two convolutional layers, each with a $5\times 5$ convolutional operation followed by batch normalization and the rectified linear unit (ReLU). In the first layer, there are more output channels than input channels to empower the network to learn more intermediate features, and in the second layer there are exactly 3 output channels to restore the image color channel. The stride of each layer depends on the required downscaling ratio. Compared to bicubic, this learnable downscaler not only adaptively trains the filters, but also incorporates non-linear operations and high-dimensional feature projection. By denoting  all the nested operations in our downscaler as $\mathcal{E}$, we obtain a small image $\mathbf{y}$ from the original image $\mathbf{x}$ via the following:
\setlength{\abovedisplayskip}{5pt}
\setlength{\belowdisplayskip}{5pt}
\begin{equation}
\mathbf{y} = \mathcal{E}\left(\mathbf{x};\mathbf{W}_{\mathrm{E}}\right).
\label{eq:y}
\end{equation}

A significant consideration in designing this downscaler is that the generated small image should remain visually pleasant and recognizable, e.g., the information in the color channels should not be destroyed or misaligned. That is to say, if pixel values in natural images follow a distribution, then the generated small image should follow the same distribution with similar moments. Hereby, we propose a moment-matching (MM) loss as follows:
\setlength{\abovedisplayskip}{5pt}
\setlength{\belowdisplayskip}{5pt}
 \begin{equation}
 \mathcal{L}_{MM} \left(\mathbf{W}_{\mathrm{E}}\right) =  \frac{1}{3}\left\| \mathcal{\mu}\left(\mathbf{x}\right) - \mathcal{\mu}\left(\mathbf{y}\right) \right\|^2_2 + \lambda \frac{1}{3}\left\| \mathcal{\sigma}\left(\mathbf{x}\right) - \mathcal{\sigma}\left(\mathbf{y}\right) \right\|_2^2,
 \label{eq:MMloss}
 \end{equation}

\noindent where $\mathcal{\mu}(\cdot)$ and $\mathcal{\sigma}(\cdot)$ compute the first and second moments respectively of the image pixel values in each color channel, and $\lambda$ is a tunable parameter that balances the two moments. This MM loss encourages that the mean and variance (loosely approximating the distribution) in each color channel of the downscaled image stay close to the mean and variance of the original image. This has been used in other application as well in the literature, such as deep generative model learning \cite{li2015generative} and style transfer \cite{li2017demystifying}.

Please note that this downscaler is not trained independently with merely the MM loss, but incorporated into the whole architecture of ThumbNet and trained together with other components and losses. This includes the classification loss, which provides supervision to the image downscaling process, guiding it to generate a small image that is discriminative for accurate classification. Hence, we name $\mathrm{E}$ a \emph{supervised} image downscaler. Once trained, this downscaler can be used for generating small images, which are not only used as input of the inference network, but also for other classification-related tasks as well.

\subsubsection{Distillation-Boosted Supervision}

A straightforward way to train the network is to minimize the classification (CL) loss defined as follows:
\setlength{\belowdisplayskip}{5pt}
\setlength{\abovedisplayskip}{5pt}
 \begin{equation}
  \mathcal{L}_{CL} \left(\mathbf{W}_{\mathrm{E}}, \mathbf{W}_{\mathrm{S}}\right) = \mathcal{H}\left(\mathbf{b}, \, \mathcal{S}\left(\mathbf{y}; \mathbf{W}_{\mathrm{S}}\right)\right),
  \label{eq:loss_c}
 \end{equation}
 where $\mathbf{y}$ is calculated as in Eq. \eqref{eq:y}, $\mathbf{b}$ indicates the ground-truth labels, $\mathcal{S}\left(\cdot\right)$ denotes all the nested functions in the inference network $\mathrm{S}$, and $\mathcal{H}$ refers to cross entropy. This loss seeks to match the predicted label with the ground-truth label and is a typical cost function for supervised learning. However, only using this loss cannot exploit the information embedded in the well-trained model $\mathrm{T}$. To address this issue, we propose to distill the learned knowledge in $\mathrm{T}$ and transfer it to the inference network. Therefore, apart from the CL loss, we also enforce the computed probabilities of each class in the inference network to match those in the well-trained network.

Let $\mathcal{S}_0\left(\cdot\right)$ and $\mathcal{T}_0\left(\cdot\right)$ denote the deep nested functions before softmax in the inference network and the well-trained network, respectively. Then, their logits are calculated as:
\setlength{\belowdisplayskip}{5pt}
\setlength{\abovedisplayskip}{5pt}
\begin{equation}
\mathbf{a}_{\mathrm{S}} = \mathcal{S}_0\left(\mathbf{y}; \mathbf{W}_{\mathrm{S}}\right), \; \mathbf{a}_{\mathrm{T}} = \mathcal{T}_0\left(\mathbf{x}\right).
 \end{equation}
Following \cite{hinton2015distilling}, we define the knowledge-distillation (KD) loss as the cross entropy between the two softened probabilities:
\setlength{\belowdisplayskip}{5pt}
\setlength{\abovedisplayskip}{5pt}
\begin{equation}
 \mathcal{L}_{KD} \left(\mathbf{W}_{\mathrm{E}}, \mathbf{W}_{\mathrm{S}}\right)  = \mathcal{H}\left( \textrm{softmax}\left(\frac{\mathbf{a}_S}{\tau}\right), \, \textrm{softmax}\left(\frac{\mathbf{a}_T}{\tau}\right) \right),
 \label{eq:loss_kd}
\end{equation}
where $\tau$ is the temperature to soften the class probabilities and is usually greater than 1. With the aid of this KD loss, the supervised training process of ThumbNet can benefit from the well-trained model to learn finer discriminative structures, so we call it distillation-boosted supervision.

\subsubsection{Feature-Mapping Regularization}

It is widely observed that unsupervised pre-training, e.g., using an auto-encoder, can help with supervised learning tasks as a form of regularization \cite{erhan2010does}. Inspired by this, we design the feature-mapping (FM) regularization to pre-train ThumbNet.

\begin{figure}[t]
\begin{center}
\includegraphics[width=0.45\textwidth]{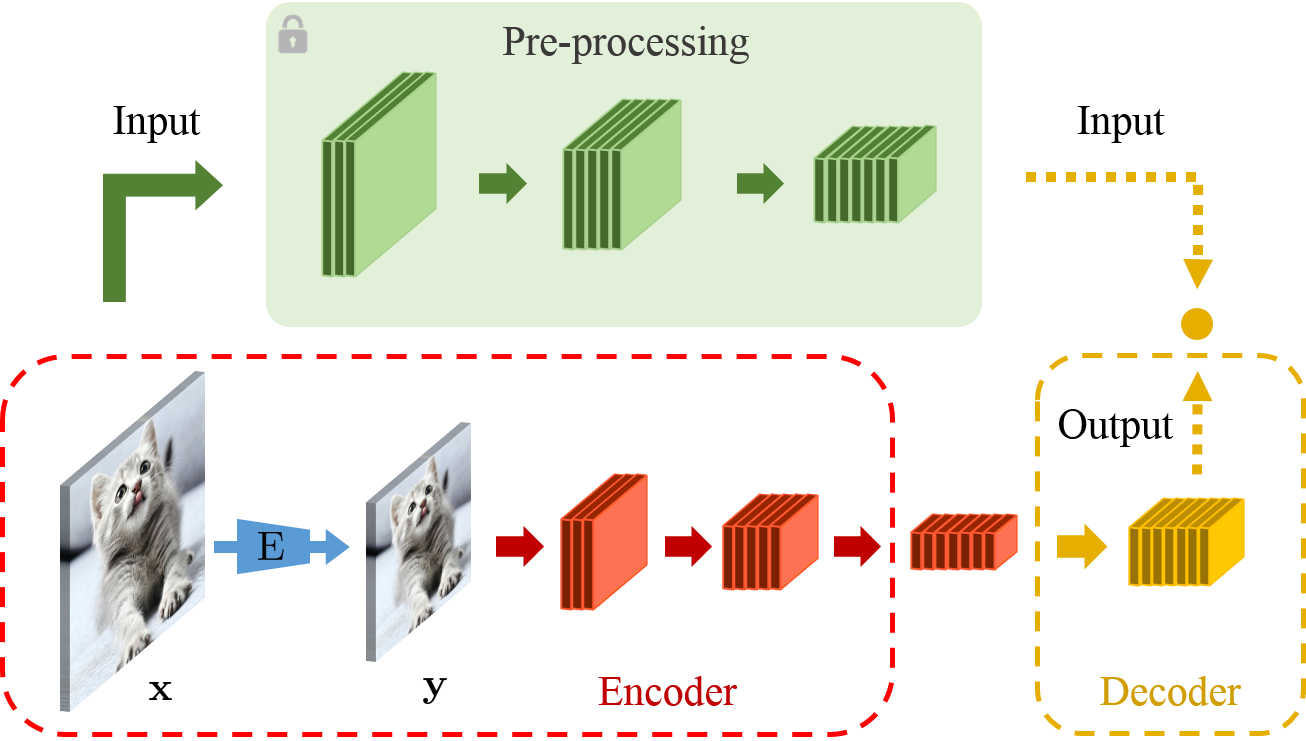}
\end{center}
\caption{\small{\textbf{Feature-mapping regularization}. This re-illustrates the left segment of ThumbNet along with the FM loss, which is essentially an auto-encoder. The layers in the red dashed box constitute the encoder, the layers in the yellow dashed box constitute the decoder, and the feature map in the middle is learned latent representation. The shaded area can be viewed as a pre-processing for the input. The yellow solid arrow represents deconvolutional layers and the yellow block is the upscaled feature map.}}
\label{fig:feature_mapping}
\end{figure}

In Fig. \ref{fig:architecture_proposed}, ThumbNet is partitioned into two segments by the FM loss (the yellow dot). In order to give a clearer sense of the rationale, we re-illustrate the left segment of ThumbNet along with the FM loss from a different point of view in Fig. \ref{fig:feature_mapping} (note that the MM loss is left out and the deconvolution in the FM loss is unrolled for the sake of clarity).  We can see that it is analogous to an auto-encoder, which is trained by minimizing the difference between the pre-processed input and the output.

In the decoder, each deconvolutional layer has a stride size 2 and the number of layers is determined by the downscaling ratio. The decoder does not change the number of channels but upscales the feature map in $\mathrm{S}$ to match the spatial size of the corresponding feature map in $\mathrm{T}$. We compute their mean square error as the FM loss:

\small
\begin{equation}
\mathcal{L}_{FM} \left(\mathbf{W}_{\mathrm{E}}, \mathbf{W}_{\mathrm{S}_l}, \mathbf{W}_\mathrm{D}\right) = \frac{1}{2N}\left\|\mathcal{T}_l \left(\mathbf{x}\right) - \mathcal{D}\left( \mathcal{S}_l\left( \mathbf{y}; \mathbf{W}_{\mathrm{S}_l}\right); \mathbf{W}_{\mathrm{D}} \right)\right\|_2^2,
\end{equation}
\normalsize
\noindent where $N$ is the product of all dimensions of the intermediate feature map in $\mathrm{T}$. $\mathcal{T}_l$ represents the nested functions in the left segment of the original network, $\mathcal{S}_l$  represents the nested functions in the left segment of the inference network, $\mathcal{D}$ represents operations in the deconvolutional layers, and $\mathbf{W}_{\mathrm{S}_l}$  and $\mathbf{W}_{\mathrm{D}}$ refer to the parameters in the left segment of $\mathrm{S}$ and in the deconvolutional layers, respectively.

\subsection{Training Details} \label{sec:training}

We summarize the training strategy of ThumbNet in Algorithm \ref{alg}. In Algorithm \ref{alg}, the $\mathcal{R}$ terms refer to ${l}_2$ regularization of the corresponding parameters, and $\theta, \alpha, \beta$ are the tradeoff weights. In Line 2, we perform an unsupervised pre-training by minimizing the MM loss and the FM loss, and obtain the parameters $\mathbf{W}^\ast_{\mathrm{E}}$ and $\mathbf{W}^\ast_{\mathrm{S}_l}$. In Line 3, we perform the distillation-boosted supervised learning by  minimizing the KD loss and CL loss, using the values $\mathbf{W}^\ast_{\mathrm{E}}$ and $\mathbf{W}^\ast_{\mathrm{S}_l}$ as initialization for the corresponding parameters. In this step, we train the whole ThumbNet end-to-end to learn the optimal values for the parameters $\mathbf{W}^\ast_{\mathrm{E}}$ and $\mathbf{W}^\ast_{\mathrm{S}}$. Note that the parameters $\mathbf{W}^\ast_{\mathrm{E}}$ and $\mathbf{W}^\ast_{\mathrm{S}_l}$ are already pre-trained via Line 2, so they are only finetuned with a small learning rate. In contrast, we use a relatively large learning rate for the untrained parameters $\mathbf{W}_{\mathrm{S}_r}$ in the right segment.

We set the hyper-parameters in ThumbNet as follows. We use a starting learning rate 0.1, and divide it by 10 when the loss plateaus. We use a momentum 0.9 for the optimizer and the weight decay $\theta$ is $0.0001$. The parameters $\alpha$ and $\beta$  in Algorithm \ref{alg} are 1.0 and 0.5, respectively; $\tau$ in Eq. \eqref{eq:loss_kd} is 2; $\lambda$ in Eq. \eqref{eq:MMloss} is 0.1. For finetuning the pre-trained parameters $\mathbf{W}_{\mathrm{E}}$ and $\mathbf{W}_{\mathrm{S}_l}$,  the learning rate is set to $0.01$ times that of the other parameters, meaning their learning rate starts from $0.001$ and is decreased by 10 in the same fashion.


\section{Experiments}\label{sec:experiments}

In this section, we demonstrate the performance of the inference network $\mathrm{S}$ obtained via ThumbNet in terms of classification accuracy and resource requirements. We also provide experiments to verify that the learned downscaler has generic applicability to other classification-related tasks as well.

\subsection{Ablation Study}

In order to study the effectiveness of the proposed techniques in ThumbNet, we ablate each technique individually and test the performance of the reduced version of ThumbNet. We provide experimental results on the object recognition and scene recognition tasks using different backbone networks (e.g., Resnet-50, VGG-11).

\subsubsection{Baseline and Comparative Methods}

\begin{table}[htbp]
\centering
\caption{\small{\textbf{Configuration of different methods}. (a) is the baseline, (b) does not use any of the three techniques, (c)-(e) contain one or two, and (f) consists of all of them. }}
\label{table:test_cases}
\def\arraystretch{1.0}
\setlength{\tabcolsep}{8pt}
\small
\begin{tabular}{l|c|c|c}
\hline

{Methods}         &SD	            &KD	            &FM	                    \\
\hline
(a) Original / Direct	                                    &$-$            &$-$	        &$-$	                \\
(b) Bicubic downscaler	                                &\texttimes	    &\texttimes	    &\texttimes             \\
(c) Supervised downscaler                               &\checkmark	    &\texttimes	    &\texttimes	            \\
(d) Bicubic + distillation                              &\texttimes	    &\checkmark	    &\texttimes             \\
(e) Supervised + distillation                           &\checkmark	    &\checkmark	    &\texttimes             \\
(f) \textbf{ThumbNet}                                   &\checkmark     &\checkmark	    &\checkmark             \\
\hline
\end{tabular}
\end{table}

We implement six variations of ThumbNet for training a backbone network, which are introduced in the following.

\vspace{3pt}\noindent\textbf{(a) Original / Direct.} `Original' refers to the network model trained on the original-size images. The testing performance of the model on an original-size image is the upperbound baseline for all the comparative methods. For networks like Resnet, which use global average pooling instead of fully-connected layers at the end, the `Original' model can be directly used to test on a small image without altering the network structure. We refer to the case of directly using the `Original' model for inference on small images without re-training as `Direct', which is the lowerbound baseline for all the methods.

\vspace{3pt}\noindent\textbf{(b) Bicubic downscaler.}  This trains the network from scratch on images that are downscaled with the bicubic method.

\vspace{3pt}\noindent\textbf{(c) Supervised downscaler.} This trains the network from scratch on small images that are downscaled with the supervised downscaler in ThumbNet. The downscaler and the network are trained jointly end-to-end based on the MM loss and the CL loss.

\vspace{3pt}\noindent\textbf{(d) Bicubic + distillation.} This trains the network on bicubic-downscaled small images with the aid of distillation from the `Original' model. The network is trained based on the KD loss and the CL loss.

\vspace{3pt}\noindent\textbf{(e) Supervised +  distillation.} This trains the network on supervised-downscaled small images with the aid of distillation from the `Original' model. The supervised downscaler and the network are trained jointly end-to-end based on the MM loss, the KD loss and the CL loss.

\vspace{3pt}\noindent\textbf{(f) ThumbNet.} This is the full configuration of our proposed method, which is trained on supervised-downscaled small images with the aid of distillation from the `Original' models as well as feature mapping regularization. It is trained based on the four losses as described in Section \ref{sec:training}.

In Table \ref{table:test_cases}, we demonstrate the configuration of each method with respect to the three techniques used in ThumbNet: supervised image downscaling (SD), knowledge-distillation boosted supervision (KD) and feature mapping regularization (FM). The hyper-parameters in all the methods are the same, as specified in Section \ref{sec:training}.

\begin{table}[htbp]
\centering
\caption{\small{\textbf{Error rates (\%) for scene recognition on Places36}. ThumbNet retains the accuracy of the original-input network when downscaling the image 16 times. }}
\label{table:accuracy_places}
\def\arraystretch{1.0}
\setlength{\tabcolsep}{5pt}
\small
\begin{tabular}{l|c|c|c|c}
\hline
     & \multicolumn{2}{c|}{Resnet-18} & \multicolumn{2}{c}{VGG-11} \\ 
      \hline
{Methods}              & Top-1            & Top-5          & Top-1         & Top-5          \\  
\hline
(a) Original               & 21.11            & 3.28           & 19.75          & 3.61          \\
(a) Direct              & 66.94            & 33.97          & /              & /             \\
(b) Bicubic downscaler               & 32.08            & 7.83           & 27.83          & 10.17         \\
(c) Supervised downscaler           & 25.94            & 5.31           & 24.28          & 6.58          \\
(d) Bicubic+distillation               & 26.00            & 4.47           & 22.69          & 4.92          \\
(e) Supervised+distillation               & 24.33            & 3.94           & \textbf{21.31}     & 3.94          \\
(f) \textbf{ThumbNet}      & \textbf{22.78}   & \textbf{3.69}  & 21.58          & \textbf{3.72} \\
\hline
\end{tabular}
\end{table}

\subsubsection{Object Recognition on ImageNet Dataset} \label{sec:performance_imagenet}

We evaluate the performance of our proposed ThumbNet on the task of object recognition with the benchmark dataset ILSVRC 2012 \cite{imagenet_cvpr09}, which consists of over one million training images drawn from 1000 categories. Besides using the full dataset (referred to as ImagenetFull), we also form a smaller new dataset by randomly selecting 100 categories from ILSVRC 2012 and refer to it as Imagenet100 (the categories in Imagenet100 are given in the \textbf{supplementary material}). With Imagenet100, we can efficiently compare all the methods on a variety of backbone networks. In addition, by partitioning ILSVRC 2012 into two parts (Imagenet100 and the rest 900 categories referred to as Imagenet900), we can also evaluate the downsampler on unseen data categories (see Section \ref{sec:eva_downscaler} for details). For backbone networks, we consider various architectures (ResNet \cite{he2016deep} and VGG \cite{simonyan2014very}) and various depths (from 11 layers to 50 layers).

In Table \ref{table:accuracy_imagenet100}, we demonstrate the performance of all the methods with four different backbone networks in terms of top-1 and top-5 error rates on the validation data. The input image size of  `Original' is $224\times 224$ and the input image sizes of the other methods are $112\times112$, meaning that in this experiment, the image downscaling ratio is $4:1$. Thus, compared to `Original', our ThumbNet only uses $1/4$ computation and memory, which will be detailed in Section \ref{sec:resource}, and it also preserves the accuracy of the original models. `Direct' is a baseline of inference on small images, compared to which our ThumbNet improves by large margins for all the different backbones. Moreover, by comparing different pairs of methods, we can obviously observe the contribution of each technique. By comparing (c) to (b) or comparing (e) to (d), we can see the benefits of supervised image downscaling. By comparing (e) to (c), we can see the benefits of distillation-boosted supervision. By comparing (f) to (e), we can see that the benefits of feature-mapping regularization. (f) ThumbNet always shows the lowest error rates.

\begin{table*}[htbp]
\centering
\caption{\small{\textbf{Error rates (\%) for object recognition on Imagenet}. By comparing (c) to (b) or comparing (e) to (d), we can see the benefits of supervised image downscaling. By comparing (e) to (c), we can see the benefits of distillation-boosted supervision. By comparing (f) to (e), we can see that the benefits of feature-mapping regularization. (f) ThumbNet always shows the lowest error rates.}}
\label{table:accuracy_imagenet100}
\def\arraystretch{1.0}
\setlength{\tabcolsep}{6pt}
\small
\begin{tabular}{l|c|c|c|c|c|c|c|c|c|c|c|c|c|c}
\hline
    & \multicolumn{8}{c|}{Imagenet100}  & \multicolumn{6}{c}{ImagenetFull} \\
\hline
   & \multicolumn{2}{c|}{VGG-11}  & \multicolumn{2}{c|}{Resnet-18}   & \multicolumn{2}{c|}{Resnet-34}  & \multicolumn{2}{c|}{Resnet-50} & \multicolumn{2}{c|}{Resnet-18}  &  \multicolumn{2}{c|}{Resnet-34} & \multicolumn{2}{c}{Resnet-50 } \\ 
      \hline
{}   & Top-1     & Top-5   & Top-1   & Top-5  & Top-1     & Top-5   & Top-1   & Top-5 & Top-1   & Top-5 & Top-1   & Top-5  & Top-1  & Top-5   \\  
\hline
(a) Original         & 13.64    & 4.36  & 17.54   & 4.98   &15.06 &4.56  & 12.72 & 3.50    & 29.71  & 10.46 & 26.19  & 8.55  & 23.59  & 6.90      \\
(a) Direct           & /    & /  & 37.26  & 16.94  &34.48 &14.32  & 29.42  & 11.80  & 50.74    & 26.15 & 45.26  & 21.67  & 38.74  & 16.72     \\
(b) Bicub. downr.     & 18.20  & 6.60  & 23.98   & 9.04  &22.42 &8.16  & 19.04  & 6.62   & 36.18 & 15.18  & 32.45  & 12.35  & 28.44  & 9.99 \\
(c) Super. downr.      & 16.14 & 5.10  & 19.70  & 7.06   &18.44 &6.16  & 16.84  & 5.66  & 34.87   & 13.98 & 31.56  & 11.88  & 27.18   & 9.24      \\
(d) Bicub. + dist.    & 17.14 & 5.84  & 20.34  & 6.64  &18.28 &5.84 & 14.96   & 4.48  & 34.76  & 14.02 & 30.60  & 10.81  & 26.28  & 8.39           \\
(e) Super. + dist.   & 17.00 & 5.78  & 17.44  & 5.16  &15.46  &4.62  & 15.12   & 3.96  & 33.02  & 12.54 & 29.18  & 10.35  & 26.13  & 8.25          \\
\textbf{(f) ThumbNet}    & \textbf{15.72}  &\textbf{4.96}   & \textbf{17.32}   & \textbf{4.98}   &\textbf{15.30} &\textbf{4.58} & \textbf{13.96}    & \textbf{3.82}  & \textbf{32.26}    & \textbf{12.13}  & \textbf{28.74}     & \textbf{9.93}      & \textbf{26.02}     & \textbf{8.25}\\
\hline
\end{tabular}
\end{table*}

\subsubsection{Scene Recognition on Places Dataset} \label{sec:performance_places}

We also apply our proposed ThumbNet to the task of scene recognition using the benchmark dataset Places365-Standard \cite{zhou2017places}. This dataset consists of 1.8 million training images from 365 scene categories. We randomly select 36 categories from Places365-Standard as our new dataset Places36 (the chosen categories are given in the \textbf{supplementary material}).

In Table \ref{table:accuracy_places}, we report the error rates on the Places36 validation dataset using two backbone networks Resnet-18 and VGG-11. The input image size of  `Original' is $224\times 224$ and the input image sizes of the other methods are $56\times56$, meaning that in this experiment, the image downscaling ratio is $16:1$. Thus, compared to `Original', our ThumbNet only uses $1/16$ computation and memory, which will be detailed in Section \ref{sec:resource}. In terms of recognition accuracy, ThumbNet nearly preserves the accuracy of the original models, where the top-5 accuracy drops by only 0.41\% for Resnet-18 and by only 0.11\% for VGG-11. Compared to `Direct', our ThumbNet improves significantly, by 44.16\% for top-1 accuracy and 30.28\% for top-5 accuracy.

\subsubsection{Resource Consumption} \label{sec:resource}

\begin{figure}[t]
\center
\includegraphics[width=0.98\linewidth]{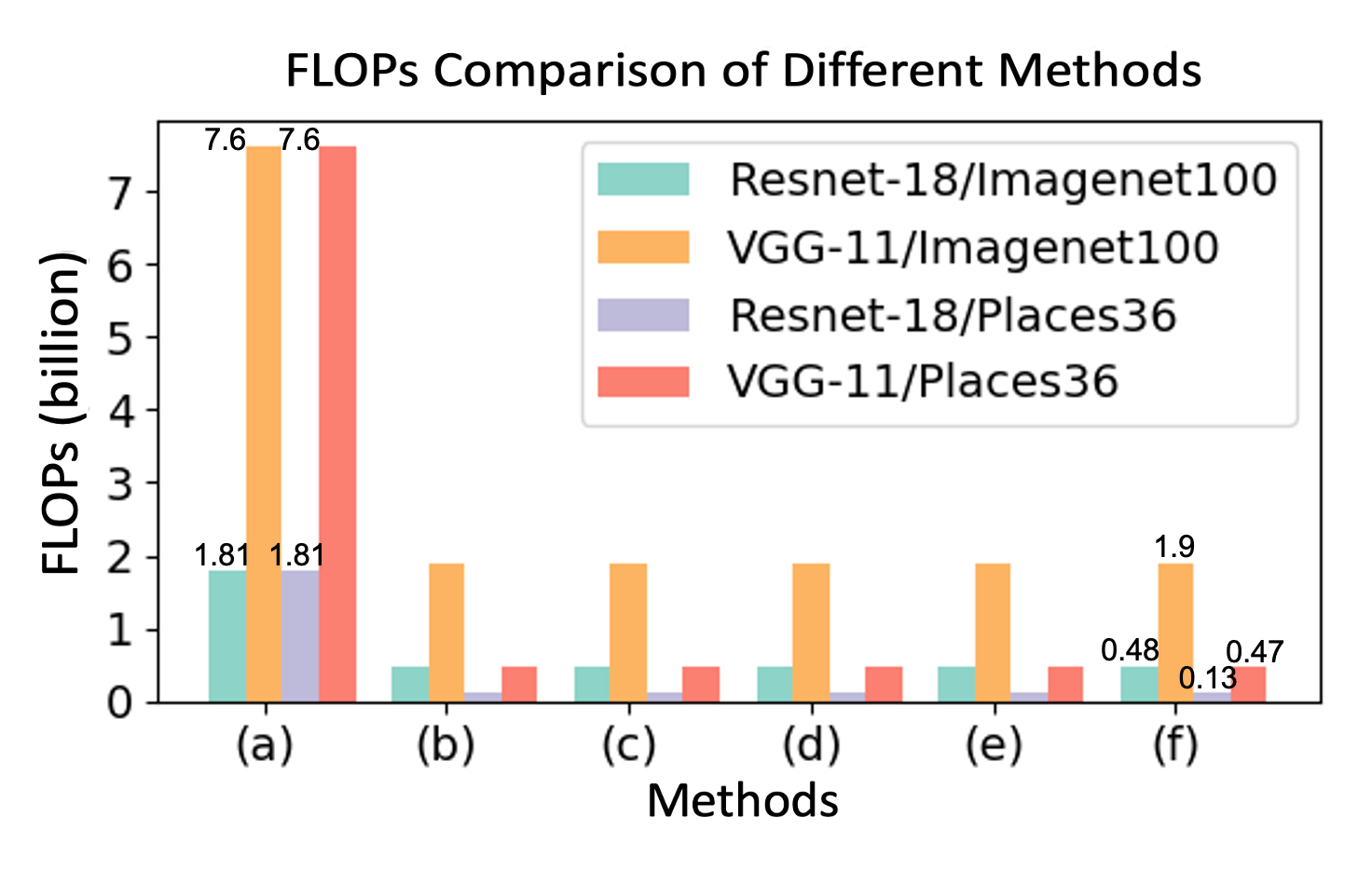}
\caption{\small{\textbf{Computation comparison of different methods}. For object recognition on Imagenet100, ThumbNet uses $1/4$ FLOPs compared to `Original'. For scene recognition on Places36, ThumbNet uses only $1/16$ FLOPs compared to `Original'. }}
\label{fig:hist}
\end{figure}

To evaluate the test-time resource consumption of the networks, we measure their number of FLoat point OPerations (FLOPs) and memory consumption of their feature maps. Fig. \ref{fig:hist} plots the number of FLOPs required by each method to classify one image for object recognition on Imagenet100 and scene recognition on Places36 with the two backbone networks Resnet-18 and VGG-11. For the task of object recognition, in which the images are downscaled $4$ times, the small-input networks use only $1/4$ FLOPs compared to the original models. For the task of scene recognition, in which the images are downscaled $16$ times, the small-input networks use only $1/16$ FLOPs compared to the original models. Similar to the reduction of computation,  the memory consumption of the 1/4-input models is about $1/4$ of the original models, and  the memory consumption of the 1/16-input models is about $1/16$ of the original models.

\begin{table*}[!h!b!t!p]
\centering
\caption{\small{\textbf{Accuracy and efficiency comparison with state-of-the-art}. B: billion; M: million; MB: MegaBytes. Regarding efficiency, ThumbNet has an obvious advantage over LWAE in terms of all metrics and for all tasks, i.e.,  ThumbNet requires fewer FLOPs, fewer network parameters, less memory feature maps, and less storage for images. Regarding classification accuracy, ThumbNet has lower top-1 and top-5 errors in the first two tasks. For the third task, ThumbNet has an obviously lower top-5 error and roughly the same top-1 error as LWAE. For the fourth task, LWAE is slightly better than ThumbNet at the cost of higher computational complexity and memory usage.}}
\label{table:compare_stateofart}
\def\arraystretch{1.0}
\setlength{\tabcolsep}{5pt}
\small
\begin{tabular}{l|c|c|c|c|c|c|c|c|c|c|c}
\hline
 \multirow{2}{*}{Tasks} &\multirow{2}{*} {{Methods}}      & \multicolumn{2}{c|}{Error rates}   & \multicolumn{2}{c|}{FLOPs} & \multicolumn{2}{c|}{Parameters} & \multicolumn{2}{c|}{Feature Memory} & \multicolumn{2}{c}{Image Storage}  \\
        \cline{3-12}
& & Top-1  & Top-5  & \# (B)  & $\downarrow$ rate & \# (M)  & $\downarrow$ rate & Size (MB)   & $\downarrow$ rate & Size (MB) &$\downarrow$ rate   \\
        \hline
Imagenet100&VGG-orig   & 13.64\%  & 4.36\%  &243.37  & 1$\times$         &129.18   &1$\times$ 	  &2118.36 &1$\times$   &4.82    &1$\times$	  	   \\
/VGG-11&LWAE \cite{chen2018learning}    & 17.98\%  & 6.42\%  &65.61  & 3.71$\times$    &67.78    &1.91$\times$ &668.94  &3.17$\times$   &2.41  	&2$\times$	 	   \\
&\textbf{ThumbNet} & \textbf{15.72\%} &\textbf{4.96\%}  &\textbf{61.04} &\textbf{3.99$\times$}	&\textbf{45.29}   &\textbf{2.85$\times$} &\textbf{530.98} 	&\textbf{3.99$\times$}   &\textbf{1.20} 	&\textbf{4$\times$}   \\
\hline
\hline
Imagenet100&Resnet-orig      & 17.54\%  &4.98\%  &58.04	&1$\times$	&11.22 	&1$\times$ 	&658.40  	&1$\times$ 	&4.82    &1$\times$  	   \\
/Resnet-18&LWAE \cite{chen2018learning}   & 21.06\%   & 6.90\%  &17.37	&3.34$\times$ 	&11.94	&0.94$\times$ 	&212.23 	&3.10$\times$ 	&2.41  &2$\times$  \\
&\textbf{ThumbNet} & \textbf{17.32\%} &\textbf{4.98\%}	 &\textbf{15.52} &\textbf{3.74$\times$} 	&\textbf{11.22}	&\textbf{1$\times$ } &\textbf{166.88}	&\textbf{3.95$\times$} 	&\textbf{1.20}  &\textbf{4$\times$} \\
\hline
\hline
Places36&VGG-orig   &19.75\% &3.61\%	&243.36	&1$\times$ 	&128.91	&1$\times$ 	&2118.33 	&1$\times$ 	&4.82 	&1$\times$     \\
/VGG-11&LWAE \cite{chen2018learning}    &\textbf{21.53\%} &4.58\%	&16.58	 &14.67$\times$ &56.50	 &2.28$\times$ 	&168.95 	&12.54$\times$ 	&0.60 	&8$\times$ \\
&\textbf{ThumbNet}  &21.58\%	&\textbf{3.72\%}	&\textbf{15.09}	&\textbf{16.13$\times$} 	&\textbf{28.25}	&\textbf{4.56$\times$} &\textbf{133.17}  &\textbf{15.91$\times$} &\textbf{0.30} &\textbf{16$\times$}  \\
\hline
\hline
Places36&Resnet-orig   &21.11\%	&3.28\%	&58.03	&1$\times$ &11.19	&1$\times$ 	&658.38   &1$\times$  &4.82  	&1$\times$  \\
/Resnet-18 &LWAE \cite{chen2018learning}    &\textbf{22.39\%}	&\textbf{3.06\%}	&4.48	&12.96$\times$ 	&11.91 &0.94$\times$ 	&54.51  	&12.08$\times$ 	&0.60   &8$\times$ \\
&\textbf{ThumbNet} &22.78\%	&3.69\%	&\textbf{4.13}	&\textbf{14.05$\times$} 	&\textbf{11.19} &\textbf{1$\times$} 	&\textbf{42.88} 	&\textbf{15.35$\times$} 	&\textbf{0.30}  	&\textbf{16$\times$ } \\
\hline
\end{tabular}
\end{table*}
\subsection{Comparison to State-of-the-Art Methods}

\subsubsection{Comparison with LWAE}

We compare our ThumbNet to LWAE \cite{chen2018learning} in terms of classification accuracy and resource requirements. Considering that the trained models of LWAE provided by the authors use different backbone networks on different datasets from ours, we re-implement LWAE in Tensorflow \cite{abadi2016tensorflow} (the same as our ThumbNet) with the same backbone networks on the same datasets as ours for fair comparison. We follow the instructions in the paper for setting the hyper-parameters and training LWAE. To evaluate both methods, we test their inference accuracy in terms of top-1 and top-5 error rates, and their inference efficiency in terms of number of FLOPs, number of network parameters, memory consumption of feature maps, and storage requirements for input images.

Table \ref{table:compare_stateofart} demonstrates the results of testing a batch of $224\times224$ color images using the two methods as well as the benchmark networks for the four tasks: object recognition on Imagenet100 with VGG-11 and Resnet-18, scene recognition on Places36 with VGG-11 and Resnet-18. For the object recognition tasks, the images are downscaled by 4 in LWAE and ThumbNet; and for the scene recognition tasks, the images are downscaled by 16 in LWAE and ThumbNet. The batch size in these experiments is set to 32.

Seen from Table \ref{table:compare_stateofart}, ThumbNet has an obvious advantage over LWAE in terms of efficiency owing to its one single input image and one single network for inference. ThumbNet only computes one network, whereas LWAE has to compute an extra branch for fusing the high-frequency image. Therefore, when downscaling an image by the same ratio, ThumbNet uses fewer FLOPs, has fewer network parameters, requires less memory for the intermediate feature maps, and stores only one small image (as compared to two images for LWAE). Regarding classification accuracy, ThumbNet has lower top-1 and top-5 errors in the first two tasks. For the third task, ThumbNet has an obviously lower top-5 error and roughly the same top-1 error as LWAE. For the fourth task, LWAE is slightly better than ThumbNet at the cost of higher computational complexity and memory usage. 

\subsubsection{Comparison with representative network acceleration / compression methods}

Table. \ref{table:compare_thinet} compares ThumbNet to representative network acceleration/compression methods in recent literature on ImagenetFull/Resnet-50. The results of the comparative methods and their respective baselines are directly taken from the published papers.
We can see that the methods SSS \cite{Huang2018DataDrivenSS}, SFP \cite{He2018SoftFP} and CP \cite{He2017ChannelPF} can only speed up by up to 2 times, which is much lower than ThumbNet. The methods ThiNet \cite{Luo2017ThiNetAF}, DCP \cite{Zhuang2018DiscriminationawareCP} and Slimmable \cite{Yu2019SlimmableNN} have similar speedup ratios (still lower than) to ThumbNet, but they have obviously higher increase in error rates.  Compared to these methods, ThumbNet has obviously lower increase in error rates with even higher acceleration ratios. Moreover, since ThumbNet is orthogonal to conventional acceleration/compression methods, it can be used in conjunction with them for further speedup.

\begin{table}[htbp]
\centering
\caption{{ThumbNet compared with representative network compression methods on Imagenet}}  
\label{table:compare_thinet}
\small
\def\arraystretch{0.98}
\setlength{\tabcolsep}{3pt}
\begin{tabular}{l|c|c|c|c|c|c}
\hline
    & \multicolumn{2}{c|}{{Top-1 error}} & \multicolumn{2}{c|}{{Top-5 error}} & \multicolumn{2}{c}{{FLOPs}} \\
\hline
{Methods}         & Err. \%     & $\uparrow$      & Err. \%    & $\uparrow$     &\# B      & $\downarrow$   \\  
\hline
$\textrm{Res50}_{\textrm{SSS}}$	   & 23.88		& 0/0\%    &7.14		& 0/0\%     & 4.09	&$1\times$     \\
SSS	\cite{Huang2018DataDrivenSS}           & 28.18	    & 4.3/18\%	         & 9.21     &2.07/29\%	         &2.32	&$1.76\times$ \\
\hline
$\textrm{Res50}_{\textrm{SFP}}$	  & 23.85		& 0/0\%       & 7.13	& 0/0\%  & --   & $1\times$   \\
SFP \cite{He2018SoftFP}	          & 25.39	    & 1.54/6\%	&7.94	&0.81/11\%		& --    & $1.72\times$      \\
\hline
$\textrm{Res50}_{\textrm{CP}}$	  & --   & --  &7.8    &0/0\%  & --  &$1\times$   \\			
CP \cite{He2017ChannelPF}			   &--   &--   &9.2	   &1.4/18\%		& --   &$2\times$    \\
\hline\hline
$\textrm{Resnet50}_{\textrm{Thi}}$     & 27.12         & 0/0\%           & 8.86          & 0/0\%         & 7.72     & $1\times$     \\
ThiNet \cite{Luo2017ThiNetAF}            & 31.58         &4.46/16\%      & 11.70         & 2.84/32\%   & 2.20     & $3.51\times$       \\
\hline
$\textrm{Resnet50}_{\textrm{DCP}}$	& 23.99       &0/0\%       &7.07   &0/0\%  & --  & $1\times$  \\		
DCP \cite{Zhuang2018DiscriminationawareCP}	        &27.25	     &3.26/14\%	 &8.87	  &1.8/25\%		& --        &$3.33\times$      \\
\hline
$\textrm{Resnet50}_{\textrm{Slim}}$	&23.9		&0/0\%    &--   &--   &4.1	& $1\times$  \\
Slimmable \cite{Yu2019SlimmableNN}	&27.9	    &4.00/17\%	     &--   &--		&1.1	&$3.73\times$     \\
\hline
$\textrm{Resnet50}_{\textrm{Thumb}}$    & 23.59         & 0/0\%           & 6.90          & 0/0\%         & 3.86      & $1\times$    \\
\textbf{ThumbNet}          & 26.02         & \textbf{2.43/10\%}     & 8.25          & \textbf{1.35/19\%}   & 1.02      & $\textbf{3.78}\times$ \\
\hline
\end{tabular}
\end{table}

\begin{figure*}[htbp]
\center
\footnotesize
\begin{subfigure}[b]{0.16\textwidth}
    \centering
    \includegraphics[width=\textwidth]{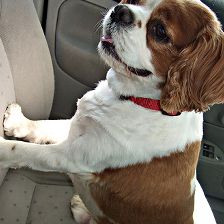}
    {Original}
\end{subfigure}
\begin{subfigure}[b]{0.16\textwidth}
    \centering
    \includegraphics[width=\textwidth]{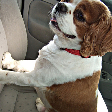}
    {Bicubic}
\end{subfigure}
\begin{subfigure}[b]{0.16\textwidth}
    \centering
    \includegraphics[width=\textwidth]{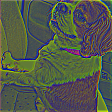}
    {Super. downscaler}
\end{subfigure}
\begin{subfigure}[b]{0.16\textwidth}
    \centering
    \includegraphics[width=\textwidth]{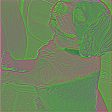}
    {Super. + distillation}
\end{subfigure}
\begin{subfigure}[b]{0.16\textwidth}
    \centering
    \includegraphics[width=\textwidth]{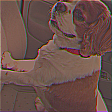}
    {ThumbNet}
\end{subfigure}
\begin{subfigure}[b]{0.16\textwidth}
    \centering
    \includegraphics[width=\textwidth]{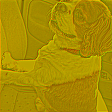}
    {W/o MM}
\end{subfigure}
\vspace{0.1em}
\caption{\small{\textbf{Visual comparison of the thumbnail images generated by different downscalers}. `Original' is of size $224\times 224$; the others are $112\times112$.  `W/o MM' means ThumbNet without the MM loss.}}
\label{fig:visual}
\end{figure*}

\subsection{Evaluation of the Supervised Downscaler} \label{sec:eva_downscaler}

\subsubsection{Visual Comparison of Downscaled Images}

In Fig. \ref{fig:visual}, we show examples of the generated thumbnail images by our ThumbNet downscalers compared to other different methods for the task of object recognition on Imagenet100 with Resnet-18. We can see that the images of `Supervised downscaler', `Supervised + distillation', ThumbNet, and `W/o MM' (ThumbNet without the MM loss) have noticeable edges compared to the `Bicubic' one. This is because these downscalers are trained in a supervised way with the classification loss being considered. The edge information in the downscaled images is helpful for making discriminative decisions. Also, note that the thumbnail image of ThumbNet is natural in color owing to the MM loss, whereas the image of `W/o MM' is obviously yellowish because the color channels are more easily messed up when the MM loss is not utilized. The top-1 error rates of ThumbNet and `W/o MM' are 17.32\% and 17.90\%, respectively, and their top-5 error rates are 4.98\% and 5.26\%, respectively. This shows that adding the MM loss does not deteriorate the network accuracy but produces more pleasant small images, which contain generic information that also benefits other tasks.

\subsubsection{Does the Supervised Downscaler Generalize?}

In order to verify that the ThumbNet downscaler is also useful apart from being used in the specific inference network $\mathrm{T}$, we also consider three different scenarios for applying our learned downscaler. Suppose that the downscaler is trained on the dataset $A$ with the backbone network $F$. We have a new dataset $A_{new}$ and a different network $F_{new}$. The three scenarios for applying the well-trained downscaler are as follows: (1) downscaling $A$ to generate small images to train $F_{new}$; (2) downscaling $A_{new}$ to generate small images to train $F$; and (3) downscaling $A_{new}$  to generate small images to train $F_{new}$.

Table \ref{table:downsampler} reports the results of these scenarios using the downscaler learned from the object recognition task with Resnet-18, which downscales an image by 4. In this case, the dataset $A$ is Imagenet100; the network $F$ is Resnet-18. We use Imagenet900 and Caltech256 \cite{griffin2007caltech} as the new dataset $A_{new}$, and VGG-11 as the new network $F_{new}$. The first two columns correspond to Scenario (1); the third and fourth columns correspond to Scenario (2); the last two columns correspond to Scenario (3), where we use the downscaler to generate thumbnail images for the new dataset Caltech256, and use these thumbnail images to train a different network VGG-11.

\begin{table}[htbp]
\centering
\caption{\small{\textbf{Performance of the ThumbNet downscaler on different networks and datasets}.}}
\label{table:downsampler}
\def\arraystretch{1.0}
\setlength{\tabcolsep}{5pt}
\small
\begin{tabular}{l|c|c|c|c|c|c}
\hline
 Networks/    & \multicolumn{2}{c|}{VGG-11/}   & \multicolumn{2}{c|}{Resnet-18/} & \multicolumn{2}{c}{VGG-11/} \\
 Datasets       & \multicolumn{2}{c|}{Imagenet100}   & \multicolumn{2}{c|}{Imagenet900} & \multicolumn{2}{c}{Caltech256} \\
        \hline
                     & Top-1   & Top-5      & Top-1     & Top-5    & Top-1    & Top-5          \\
        \hline
Original              & 13.64      & 4.36        & 28.80        & 10.17       & 30.68	      & 18.56            \\
Bicubic           & 18.20      & 6.60         & 35.52        & 14.56       & 32.80	      & 19.63             \\
\textbf{ThumbNet}  & \textbf{16.74}  & \textbf{6.42}  & \textbf{33.54}  & \textbf{13.22}  & \textbf{32.04}	 & \textbf{18.56}     \\
\hline
\end{tabular}
\end{table}

The first row shows the performance of the respective networks trained on the original-size images, while the second row shows their performance when trained on the small images downscaled with bicubic interpolation. The third row shows the performance of the networks trained on the small images downscaled by our ThumbNet downscaler. We can see that the third row obviously outperforms the second row in all scenarios, indicating that our supervised downscaler tends to generalize to other datasets and other network architectures. In fact, it is very promising to see that in Scenario (3) when both the dataset and the network are new to the downscaler, it can still bring about significant gains compared to the bicubic naive downscaler, leading to the same top-5 error rate as the original network.

\section{Conclusions}\label{sec:conclusions}

In this paper, we propose a unified framework ThumbNet to tackle the problem of accelerating run-time deep convolutional network from a novel perspective: downscaling the input image. Based on the fact that reducing the input image size lowers the computation and memory costs of a CNN, we seek to obtain a network that can retain original accuracy when applied on one thumbnail image. Experimental results show that, with our ThumbNet, we are able to learn a network that dramatically reduces resource requirements without compromising classification accuracy. Moreover, we have a supervised downscaler as a side product, which can be utilized for generic classification purposes, generalizing to datasets and network architectures that it was not exposed to in training. This work can be used in conjunction with other network acceleration/compression methods for further speed up without incurring additional overheads.

\bibliographystyle{ACM-Reference-Format}
\bibliography{main}


\onecolumn
\begin{center}
    \textbf{\LARGE{Supplementary Material}}
 \end{center}
 
\setcounter{section}{0}







\section{Details of the datasets Imagenet100 and Places36}
\subsection{Imagenet100}
\noindent The 100 categories in Imagenet100 for the object recognition task in Section 4.1.2 are listed as follows
\newline

\begin{xtabular}{ll}
n02808440 & \quad bathtub, bathing tub, bath, tub                                                     \\
n02100877 & \quad Irish setter, red setter                                                            \\
n02096585 & \quad Boston bull, Boston terrier                                                         \\
n03447721 & \quad gong, tam-tam                                                                       \\
n03804744 & \quad nail                                                                                \\
n03930313 & \quad picket fence, paling                                                                \\
n03908714 & \quad pencil sharpener                                                                    \\
n02097298 & \quad Scotch terrier, Scottish terrier, Scottie                                           \\
n04179913 & \quad sewing machine                                                                      \\
n02169497 & \quad leaf beetle, chrysomelid                                                            \\
n04141327 & \quad scabbard                                                                            \\
n07768694 & \quad pomegranate                                                                         \\
n03938244 & \quad pillow                                                                              \\
n04133789 & \quad sandal                                                                              \\
n04008634 & \quad projectile, missile                                                                 \\
n01632777 & \quad axolotl, mud puppy, Ambystoma mexicanum                                             \\
n02096177 & \quad cairn, cairn terrier                                                                \\
n03000134 & \quad chainlink fence                                                                     \\
n07860988 & \quad dough                                                                               \\
n03417042 & \quad garbage truck, dustcart                                                             \\
n04550184 & \quad wardrobe, closet, press                                                             \\
n04542943 & \quad waffle iron                                                                         \\
n02487347 & \quad macaque                                                                             \\
n02007558 & \quad flamingo                                                                            \\
n04443257 & \quad tobacco shop, tobacconist shop, tobacconist                                         \\
n03902125 & \quad pay-phone, pay-station                                                              \\
n04418357 & \quad theater curtain, theatre curtain                                                    \\
n02128925 & \quad jaguar, panther, Panthera onca, Felis onca                                          \\
n02101388 & \quad Brittany spaniel                                                                    \\
n02860847 & \quad bobsled, bobsleigh, bob                                                             \\
n13040303 & \quad stinkhorn, carrion fungus                                                           \\
n04355338 & \quad sundial                                                                             \\
n01774384 & \quad black widow, Latrodectus mactans                                                    \\
n03657121 & \quad lens cap, lens cover                                                                \\
n02708093 & \quad analog clock                                                                        \\
n04111531 & \quad rotisserie                                                                          \\
n01829413 & \quad hornbill                                                                            \\
n04204347 & \quad shopping cart                                                                       \\
n03792782 & \quad mountain bike, all-terrain bike, off-roader                                         \\
n02268443 & \quad dragonfly, darning needle, devil's darning needle, sewing needle,                    \\
          & \quad snake feeder, snake doctor, mosquito hawk, skeeter hawk                             \\
n03933933 & \quad pier                                                                                \\
n02879718 & \quad bow                                                                                 \\
n03770439 & \quad miniskirt, mini                                                                     \\
n03125729 & \quad cradle                                                                              \\
n03127747 & \quad crash helmet                                                                        \\
n01728920 & \quad ringneck snake, ring-necked snake, ring snake                                       \\
n03769881 & \quad minibus                                                                             \\
n04404412 & \quad television, television system                                                       \\
n01530575 & \quad brambling, Fringilla montifringilla                                                 \\
n04033995 & \quad quilt, comforter, comfort, puff                                                     \\
n02102318 & \quad cocker spaniel, English cocker spaniel, cocker                                      \\
n03658185 & \quad letter opener, paper knife, paperknife                                              \\
n01677366 & \quad common iguana, iguana, Iguana iguana                                                \\
n01930112 & \quad nematode, nematode worm, roundworm                                                  \\
n01496331 & \quad electric ray, crampfish, numbfish, torpedo                                          \\
n02219486 & \quad ant, emmet, pismire                                                                 \\
n02437312 & \quad Arabian camel, dromedary, Camelus dromedarius                                       \\
n04258138 & \quad solar dish, solar collector, solar furnace                                          \\
n04596742 & \quad wok                                                                                 \\
n02859443 & \quad boathouse                                                                           \\
n02356798 & \quad fox squirrel, eastern fox squirrel, Sciurus niger                                   \\
n02777292 & \quad balance beam, beam                                                                  \\
n12998815 & \quad agaric                                                                              \\
n02951358 & \quad canoe                                                                               \\
n03782006 & \quad monitor                                                                             \\
n03676483 & \quad lipstick, lip rouge                                                                 \\
n03532672 & \quad hook, claw                                                                          \\
n02749479 & \quad assault rifle, assault gun                                                          \\
n04325704 & \quad stole                                                                               \\
n04026417 & \quad purse                                                                               \\
n09256479 & \quad coral reef                                                                          \\
n07742313 & \quad Granny Smith                                                                        \\
n01687978 & \quad agama                                                                               \\
n02835271 & \quad bicycle-built-for-two, tandem bicycle, tandem                                       \\
n01667778 & \quad terrapin                                                                            \\
n03187595 & \quad dial telephone, dial phone                                                          \\
n02113023 & \quad Pembroke, Pembroke Welsh corgi                                                      \\
n01739381 & \quad vine snake                                                                          \\
n02120079 & \quad Arctic fox, white fox, Alopex lagopus                                               \\
n02056570 & \quad king penguin, Aptenodytes patagonica                                                \\
n04435653 & \quad tile roof                                                                           \\
n01749939 & \quad green mamba                                                                         \\
n03207941 & \quad dishwasher, dish washer, dishwashing machine                                        \\
n07831146 & \quad carbonara                                                                           \\
n04604644 & \quad worm fence, snake fence, snake-rail fence, Virginia fence                           \\
n02927161 & \quad butcher shop, meat market                                                           \\
n01630670 & \quad common newt, Triturus vulgaris                                                      \\
n03598930 & \quad jigsaw puzzle                                                                       \\
n03691459 & \quad loudspeaker, speaker, speaker unit, loudspeaker system, speaker system              \\
n02114855 & \quad coyote, prairie wolf, brush wolf, Canis latrans                                     \\
n02791270 & \quad barbershop                                                                          \\
n01484850 & \quad great white shark, white shark, man-eater, man-eating shark, Carcharodon carcharias \\
n04146614 & \quad school bus                                                                          \\
n04356056 & \quad sunglasses, dark glasses, shades                                                    \\
n02086646 & \quad Blenheim spaniel                                                                    \\
n02110627 & \quad affenpinscher, monkey pinscher, monkey dog                                          \\
n03854065 & \quad organ, pipe organ                                                                   \\
n03697007 & \quad lumbermill, sawmill                                                                 \\
n02454379 & \quad armadillo                                                                           \\
n03314780 & \quad face powder
\end{xtabular}%
\newline

\subsection{Places36}
\noindent The 36 categories in Places36 for the scene recognition tasks in Section 4.1.3 are listed as follows
\newline

\begin{xtabular}{ll}
307 & \quad /s/skyscraper                \\
29  & \quad /a/auto\_showroom            \\
195 & \quad /j/jacuzzi/indoor            \\
13  & \quad /a/archaelogical\_excavation \\
108 & \quad /c/courthouse                \\
16  & \quad /a/arena/performance         \\
1   & \quad /a/airplane\_cabin           \\
38  & \quad /b/banquet\_hall             \\
175 & \quad /h/highway                   \\
268 & \quad /p/playground                \\
129 & \quad /e/elevator/door             \\
142 & \quad /f/field\_road               \\
123 & \quad /d/doorway/outdoor           \\
11  & \quad /a/arcade                    \\
338 & \quad /t/tree\_farm                \\
151 & \quad /f/forest\_path              \\
47  & \quad /b/bazaar/outdoor            \\
140 & \quad /f/field/cultivated          \\
181 & \quad /h/hotel/outdoor             \\
107 & \quad /c/cottage                   \\
155 & \quad /g/galley                    \\
178 & \quad /h/hospital                  \\
51  & \quad /b/bedchamber                \\
4   & \quad /a/alley                     \\
141 & \quad /f/field/wild                \\
262 & \quad /p/pharmacy                  \\
297 & \quad /s/science\_museum           \\
79  & \quad /c/canal/urban               \\
254 & \quad /p/park                      \\
293 & \quad /r/runway                    \\
100 & \quad /c/computer\_room            \\
99  & \quad /c/coffee\_shop              \\
210 & \quad /l/lecture\_room             \\
362 & \quad /y/yard                      \\
45  & \quad /b/bathroom                  \\
211 & \quad /l/legislative\_chamber
\end{xtabular}%



\end{document}